# DS-Diffusion: Data Style-Guided Diffusion Model for Time-Series Generation

Mingchun Sun, Rongqiang Zhao*, Hengrui Hu, Songyu Ding, Jie Liu *Fellow, IEEE*

*Abstract*—Diffusion models are the mainstream approach for time series generation tasks. However, existing diffusion models for time series generation require retraining the diffusion backbone to introduce specific conditional guidance. There also exists a certain degree of distributional bias between the generated data and the real data, which leads to potential model biases in downstream tasks. Additionally, the complexity of diffusion models and the latent spaces leads to an uninterpretable inference process. To address these issues, we propose the data style-guided diffusion model (DS-Diffusion). In the DS-Diffusion, a diffusion framework based on style-guided kernels is developed to avoid retraining the diffusion backbone for specific conditions. The time-information based hierarchical denoising mechanism (THD) is developed to reduce the distributional bias between the generated data and the real data. Furthermore, the generated samples can clearly indicate the data style from which they originate. We conduct comprehensive evaluations using multiple public datasets to validate our approach. Experimental results show that, compared to the state-of-the-art model such as ImagenTime, the predictive score and the discriminative score decrease by 5.56% and 61.55%, respectively. The distributional bias between the generated data and the real data is further reduced, the inference process is also more interpretable. Moreover, by eliminating the need to retrain the diffusion model, the flexibility and adaptability of the model to specific conditions are also enhanced.

## I. INTRODUCTION

Data generation is crucial for reducing model overfitting and improving model performance [1]. When constructing datasets for training models, data collection faces various challenges, including high costs, time consumption, and the scarcity of target samples. Data annotation also requires specialized knowledge and involves the annotator's subjective judgment [2]. Differences in personal experience and knowledge also lead to inconsistent annotation results [3], [4]. Additionally, specific patterns reflected in time series are data styles. By data styles, we refer to trends and seasonality. The core structure of the generated data distribution is defined by the data styles, making them critical for data generation [5]. Therefore, it is essential to augment the dataset by generating data that conform to the styles of real data. The generated data could increase the quantity and diversity of the dataset, thereby improving the robustness and generalization ability of models in downstream tasks.

Existing data generation approaches are primarily categorized into data-driven and model-driven approaches. Data-driven approaches generate new samples through the transformation of real samples. The transformation includes mathematical transformations, time-frequency conversions, sequence decomposition and recombination [6], [7], [8]. While the data-level approaches offer a degree of flexibility, they struggle to capture the correlations among variables in multivariate datasets [9]. With the advancement of deep learning techniques, model-driven generation approaches have gradually emerged. Generative models learn the latent distribution of data to generate samples, capturing more complex patterns and features within the samples and among variables [10], [11]. Mainstream generative models include generative adversarial networks (GANs), variational autoencoders (VAEs), and diffusion models [12], [13]. VAEs may not adequately capture the complexity of data in latent space, and GANs often exhibit training instability which makes them prone to mode collapse [14]. In contrast, diffusion models generate samples through a process of gradually adding and removing noise. By eliminating the need for adversarial training, diffusion models avoid mode collapse, thus providing relatively stable training [15], [16]. As a result, diffusion models have become the mainstream approach widely used in generative tasks. Moreover, the data styles, which include the trend component and the seasonal component, are used as a condition to guide the training of diffusion models [17], [18]. The condition guides the diffusion model to denoise in a way that aligns more closely with the data styles. The generated data becomes more similar to the real data. So the similarity between the generated data and the real data is enhanced.

However, with the widespread use of diffusion models in unconditional time series generation tasks, several shortcomings gradually emerge. Firstly, diffusion models generate samples by learning the latent distribution of the data, which is a complex training process. For existing conditional diffusion models in unconditional time series generation tasks, generating samples that meet specific conditions requires retraining the diffusion backbone or classifiers. The complex training process limits the model's flexibility and adaptability to specific application scenarios. The inability to update or adjust the model in a timely manner based on the generated targets limits the diffusion model's capacity to fulfill the requirements for diverse sample generation. Secondly, there remains an unstable bias between the distribution of the generated data and the real data. The distributional bias results in the generated data that

*Corresponding author: Rongqiang Zhao@zhaorq@hit.edu.cn.

Mingchun Sun, Rongqiang Zhao, Hengrui Hu, Songyu Ding, and Jie Liu are with the Faculty of Computing, Harbin Institute of Technology, Harbin 150001, China.

Mingchun Sun, Rongqiang Zhao, Hengrui Hu, Songyu Ding, and Jie Liu are with the National Key Laboratory of Smart Farm Technologies and Systems, Harbin 150001, China.

do not effectively reflect the characteristics of the real data, potentially causing model bias in downstream tasks. So the accuracy and reliability of predictions are affected. Thirdly, the complexity of the latent space in diffusion models complicates the inference process, making it difficult to provide a clear explanation for the source of the generated data. Although certain feature representations of time series are used to guide the training of diffusion models, the generated data do not directly indicate their origins. The credibility of the generated data is limited.

To address these issues, we propose the **Data Style-guided Diffusion** model (**DS-Diffusion**) for time series generation tasks. In the data styles that we refer to, the trend component represents the overall direction of change in time series data over a longer time span, the seasonal component reflects the regular fluctuations exhibited by the time series within specific periods. The DS-Diffusion extracts data styles from real data to guide the inference process. The introduction of data styles does not require retraining the diffusion model, enhancing its flexibility and adaptability to specific conditions for generation. Additionally, due to the guidance of data styles, the distributional bias between the generated data and the real data is reduced, and the inference process is more interpretable. The main contributions of our work are summarized as follows:

- We propose a diffusion model framework based on style-guided kernels. The style-guided kernels modify the inference process directly. There is no need for retraining the diffusion backbone when introducing new conditions for generation. Therefore, the flexibility and adaptability of the diffusion model are enhanced.
- We propose a time-information based hierarchical denoising mechanism (THD) in the style-guided kernels. In the inference process, by guiding the denoising of different frequency components at various denoising time steps, the distributional bias between the generated data and the real data is reduced. The inference process is also more interpretable.
- We conduct extensive experiments on multiple public time series datasets. Compared to the strong baseline ImagenTime in generation, our approach achieves state-of-the-art results in the predictive score and the discriminative score. Notably, our approach obtains lower scores on metrics reflecting distributional bias.

## II. RELATED WORK

### A. Diffusion models

In recent years, diffusion models have gained widespread attention, particularly for their exceptional performance in the field of computer vision [19]. Diffusion models are initially proposed for image generation with the inception traceable to 2015 [20], where the idea revolves around the gradual addition and removal of noise to generate samples. The generation process draws inspiration from diffusion processes in physics. With the advancement of deep learning technologies, the potential of diffusion models is progressively explored. DDPM [21], DDIM [22], and score-based generative models [23] emerge as mature frameworks, demonstrating significant generative performance. Building on these diffusion models, various approaches for incorporating external conditions are developed to guide the generation of samples. The introduction of external conditions allows diffusion models to generate samples in a directed manner, progressively steering the generation process toward satisfying predefined requirements [24], [25]. The expansion of multimodal data further enriches the information available to diffusion models, enhancing their generative capabilities [26], [27]. Consequently, diffusion models exhibit robust generative power and flexibility within the field of computer vision. Additionally, a promising approach involves establishing the transformation between time series and images. The transformation allows advanced diffusion model techniques from computer vision to be applied to time series generation tasks [28], [14]. However, despite the considerable improvements in the quality and diversity of generated data due to the incorporation of external conditions and multimodal information, some distributional bias remains between the generated data and the real data. Furthermore, the inclusion of conditional information increases the complexity of the network and the noise processing, which leads to a more uninterpretable inference process.

### B. Generative models for time series

Mainstream generative models include GANs, VAEs, and diffusion models, which exhibits outstanding results across various time series datasets. However, GANs exhibit poor training stability due to the adversarial mechanism between the generator and discriminator, often leading to mode collapse [14]. VAEs have limited capacity in learning the data distribution within the latent space, resulting in insufficient modal coverage and diversity of the generated data. Guiding of prior knowledge alleviates this issue to some extent [29]. In contrast, diffusion models provide higher generative quality, stronger learning capacity, and controllability, along with a more stable training process in generative tasks [30], [31]. Consequently, diffusion models gradually replace GANs and VAEs as the mainstream models for generative tasks. Various diffusion models designed for different scenarios of time series data are being proposed. The primary direction focuses on incorporating external conditions as prior knowledge into the processes of adding and removing noise [32], [33]. Given the advantages of transformers and GRUs in processing sequential data, they are used to extract conditions from time series [34], [35]. The extracted conditions encompass sequence patterns and frequency domain information, which are applied to guide the training of diffusion models [36]. The conditions can be regarded as different data styles. Additionally, the loss function is modified to incorporate the corresponding conditions [37]. In the field of time series, among the advanced diffusion models developed, despite the promising generative capabilities of diffusion models guided by different conditions on time series datasets, the introduction of networks and strategies complicates the training process and increases computational resource consumption. Particularly, when generating data that meet specific conditions, the diffusion backbone requires retraining. This not only raises training costs but also diminishes

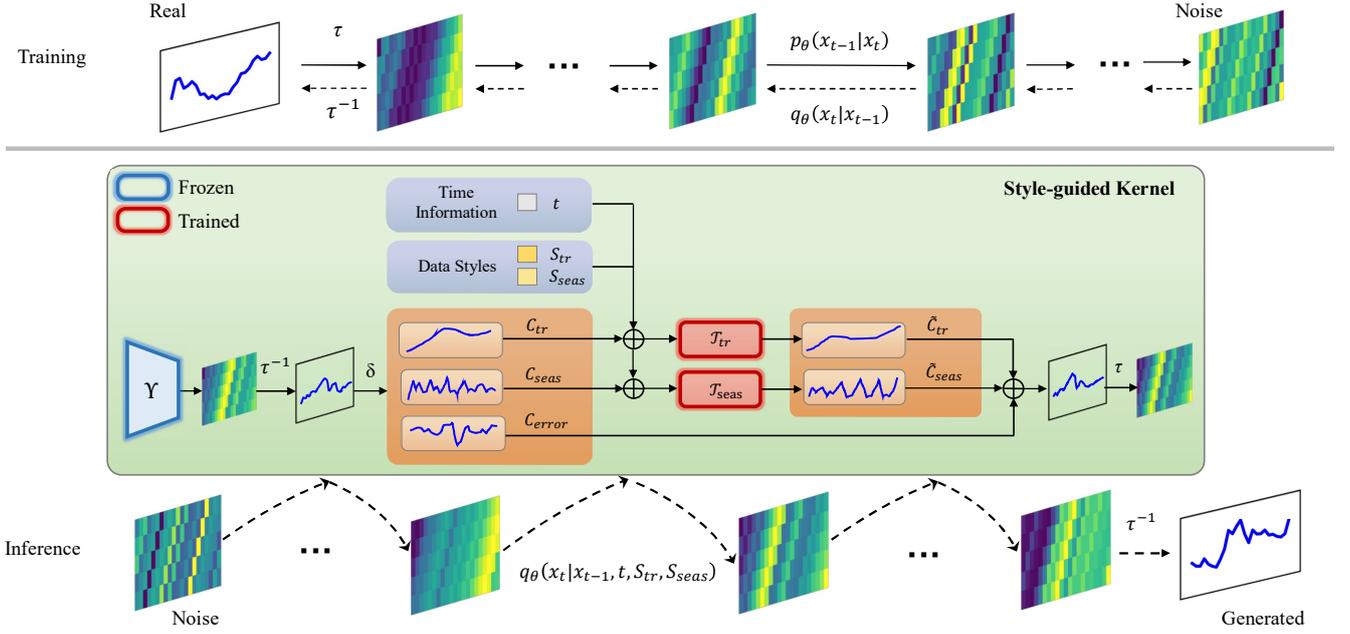

Fig. 1. Illustration of DS-Diffusion. In framework, style-guided kernels do not participate in training process of diffusion backbone but is solely involved in the inference process. To simplify, we depict the denoising of a sample during a middle denoising timestep in the style-guided kernel. Both the time information ($t$) and the data styles ($S_{tr}$, $S_{seas}$) are used to guide the denoising.

the flexibility and adaptability of diffusion models in dynamic environments.

### III. METHODOLOGY

#### A. Problem formulation

Assuming a time series dataset $\mathcal{D}^{L \times F} = \{x | x_0^{l \times F}, x_1^{l \times F}, \ldots, x_n^{l \times F}\}$, where $L$ denotes the sequence length in $\mathcal{D}^{L \times F}$, $F$ represents the dimensionality of features in $\mathcal{D}^{L \times F}$, $l$ indicates the length of each sample, and $n$ denotes the number of samples in the dataset. $x^{l \times F} \in \mathcal{D}^{L \times F}$ represents a sample in the dataset. $S = \phi(x^{l \times F})$ signifies the extracted data style from each sample, which includes the trend component $S_{tr} \in \mathbb{R}^{l \times F}$ and the seasonal component $S_{seas} \in \mathbb{R}^{l \times F}$. The critical issue is how to effectively leverage $S$ to guide the inference process. High-quality generated data should exhibit a strong statistical consistency with the real data. Therefore, if the learned data distribution $\widetilde{p}(x)$ approaches the true data distribution $p(x)$ more closely, it indicates that the quality of the generated data sampled from $\widetilde{p}(x)$ is better.

#### B. Overview

Time series exhibit various data styles in real-world scenarios across many domains. By data styles, we refer to trend components and seasonal components, which determine the richness and diversity of samples within a dataset. The core idea of the DS-Diffusion is to guide the inference process of the diffusion model using data styles derived from real samples. At different denoising time steps of the inference process, various frequency components of the generated samples receive more detailed guidance. The DS-Diffusion consists of two main components: (i) a diffusion model framework based on style-guided kernels; (ii) the THD. Apart from the trained diffusion model, the style-guided kernel is the only part of our approach that contains learnable parameters. The DS-Diffusion is illustrated in Fig. 1.

#### C. Diffusion Model Framework based on Style-guided Kernels

The ImagenTime [14] is the foundational framework, where the EDM [38] is used as the diffusion generative backbone in the training and the inference process, denoted as $\Upsilon$. In the DS-Diffusion, the inference begins with a Gaussian noise sample $x_T \sim N(0, I)$, and gradually performs denoising inference to obtain the generated sample $x_0$ through the diffusion model framework based on style-guided kernels. At each inference time step $t \in [0, T]$ of the $\Upsilon$ framework, the logarithmic likelihood gradient of the current $x_t$ is first calculated by the trained denoising model, as described in Eq. (1), where $\Psi$ represents the trained denoising model within the $\Upsilon$ framework.

$$\nabla_{x_t} \log p_t(x_t) = \Psi(x_t, t) \quad (1)$$

$\Psi$ is used to denoise and minimize the divergence from the real data distribution, typically using the variational lower bound for optimization. The loss function of $\Psi$ can be expressed in Eq. (2):

$$L = \mathbb{E}_{t, x_0, \epsilon}\left[\|\epsilon - \epsilon_\theta(x_t, t)\|^2\right] \quad (2)$$

$\epsilon_\theta(x_t, t)$ is the noise predicted by the denoising model. Next, the logarithmic likelihood gradient is used to denoise and update the current sample $x_t$, as described in Eq. (3).

$$dx_t = \left[f(t, x_t) - g(t)^2 \nabla_{x_t} \log p_t(x_t)\right] dt + g(t)dW_t \quad (3)$$

$f(t, x_t)$ represents the drift component, which guides the generated data distribution at time step $t$ to move towards the real data distribution. $g(t)$ denotes the diffusion coefficient that controls the intensity of the noise, and $dW_t$ signifies the increment of Brownian motion that reflects the generated uncertainty. In Eq. (3), $t$ denotes a continuous diffusion time variable defined in $[0, 1]$, while in practice the inference is performed in discrete steps $t \in 0, 1, \ldots, T$. Finally, $x_t$ is processed by the style-guided kernel, as indicated in Eq. (4).

$$x_{t-1} = \mathcal{K}(S, x_t), t \in 0, 1, \ldots, T \quad (4)$$

$\mathcal{K}$ denotes the style-guided kernels, $S$ represents the data styles, and $x_{t-1}$ is the final updated sample in the current denoising time step. In the inference process, the style-guided kernel leverages the data styles from real samples to guide the sample updates. Therefore, even with the introduction of different data styles as conditions, the $\Upsilon$ does not require retraining. The core of the style-guided kernel is the THD.

*D. THD*

In the data styles of time series, the trend component is considered a low-frequency component, while the seasonal component is regarded as a high-frequency component. Introducing data styles that mix different frequency components directly into the denoising process for guidance could cause interference between high-frequency and low-frequency features, thereby affecting the accurate capture and reconstruction of the real samples. Therefore, in each style-guided kernel, the THD is used to guide the denoising of different frequency components in the inference process based on data styles.

In the THD, the generated samples are composed of guided trend components, guided seasonal components, and retained residual components in each denoising time step. Given the dataset $\mathcal{D}^{L \times F}$, prior to the inference, data styles from real samples are extracted. Seasonal and trend decomposition using loess (STL) is used to extract the data styles, denoted as $\delta$, as described in Eq. (5).

$$S_{tr}, S_{seas} = \delta(x_i^{l \times F}), i \in [0, n] \quad (5)$$

The extracted data styles include the trend component $S_{tr}$ and the seasonal component $S_{seas}$, corresponding to the low-frequency and high-frequency components in the samples, respectively. The core of THD is the guidance of data styles for different frequency components at various denoising time steps. At each denoising time step, the the denoised result $x_t$ is first obtained through the trained denoising model, and then $x_t$ is transformed into the time series $x_t^{ts}$. Through the THD, $x_t^{ts}$ is decomposed into the trend component $C_{tr}$, the seasonal component $C_{seas}$, and the residual component $C_{error}$. This process is described in Eq. (6) and Eq. (7).

$$x_t^{ts} = \tau^{-1}(x_t) \quad (6)$$

$$C_{tr}, C_{seas}, C_{error} = \delta(x_t^{ts}) \quad (7)$$

$\tau^{-1}$ in Eq. (6) represents the process of transforming back to the time series using the image inverse transformation [14]. In Eq. (5), $i$ refers to the sample index in the given dataset, which is independent of the inference time step $t$ used in Eq. (7). $C_{error}$ is the noise component of the sample that is entirely random. Random noise does not carry information about the samples but affects the reconstruction of the sample, particularly affecting the high-frequency components. Therefore, $C_{error}$ is retained. In the hierarchical denoising process of the THD, $S_{tr}$ is used to guide the denoising of the low-frequency component $C_{tr}$, and $S_{seas}$ is used to guide the denoising of the high-frequency component $C_{seas}$, with $C_{error}$ not participating in the hierarchical denoising process.

The Fourier transform of the time series reveals the manner in which high-frequency and low-frequency components change during the denoising process. In the early time steps of denoising inference, low-frequency components have strong power, leading to a higher signal-to-noise ratio. Low-frequency components could be clearly identified and used to generate the outline of the samples. Due to the lower signal-to-noise ratio of high-frequency components, their attention is reduced to avoid generating erroneous information in a low signal-to-noise ratio scenario. As the inference progresses, high-frequency components gradually receive attention to enhance the details of the generated samples. As a result, during the earlier time steps of the inference process, the THD places greater emphasis on the low-frequency component. The data styles are more inclined to guide the denoising of $C_{tr}$. Furthermore, the attention to the high-frequency component is suppressed, resulting in relatively less guidance for $C_{seas}$. The structural information of the generated samples is gradually reconstructed, and the outline of the sequence is established. On the contrary, in the later time steps of the inference process, the THD increases its focus on the high-frequency component. The data styles are more inclined to guide the denoising of $C_{seas}$, while attention to $C_{tr}$ gradually decreases. Therefore, the guidance of data styles for seasonal features is enhanced, and the generated sequence outline becomes progressively refined. At the same time, attention to the low-frequency component $C_{tr}$ is gradually suppressed, ensuring that the completed reconstructed sequence outline remains unchanged.

Due to the time series learning capabilities of transformers, the transformer is the core component of the THD. At each denoising time step of the inference process, the independent transformers learns the attention to different frequency components and the guiding role of data styles. The training of the transformers involves three key steps. Firstly, the decomposed trend and seasonal components are separately combined with temporal information and data styles. The input information is then mapped to a high-dimensional space through an embedding layer. Secondly, the global dependencies within the sequence are learned using a multi-head self-attention mechanism, which models the low-frequency trend and high-frequency seasonal features independently. Thirdly, the output layer predicts the seasonal and trend components that guide the denoising process, allowing the model to iteratively reconstruct the sample structure and details over multiple time steps. Since the prediction of trend and seasonal components is a regression task, Mean Squared Error (MSE) is utilized as the loss function. Adam is employed as the optimizer. The transformers are employed to predict the guided trend component and guided seasonal component of the generated samples. Specifically, the time information $t$, $S_{tr}$ in data styles, and the low-frequency component of the current sample $C_{tr}$ are input into the transformer for the denoising of low-frequency components, which is denoted as $\mathcal{T}_{tr}$. The time information $t$, $S_{seas}$ in data styles, and the low-frequency component of the current sample $C_{seas}$ are input into the transformer for the denoising of high-frequency components, which is denoted as $\mathcal{T}_{seas}$. The trained $\mathcal{T}_{tr}$ and $\mathcal{T}_{seas}$ would be utilized at each denoising time step. $\widetilde{C}_{tr}$ and $\widetilde{C}_{seas}$ represent the denoising results of the low-frequency component and the high-frequency component. The denoising results are directly computed, as described by Eq. (8) and Eq. (9).

$$\widetilde{C}_{tr} = \mathcal{T}_{tr}(C_{tr}, S_{tr}, t) \qquad (8)$$

$$\widetilde{C}_{seas} = \mathcal{T}_{seas}(C_{seas}, S_{seas}, t) \qquad (9)$$

Then $\widetilde{C}_{tr}$ and $\widetilde{C}_{seas}$ are used, along with $C_{error}$, to synthesized a complete sample at the current time step, denoted as $\widetilde{x}_t^{ts}$. $\widetilde{x}_t^{ts}$ is transformed into the image form (denoted as $x_{t-1}$) and sent to the next denoising time step, as described by Eq. (10) and Eq. (11).

$$\widetilde{x}_t^{ts} = \widetilde{C}_{tr} + \widetilde{C}_{seas} + C_{error} \qquad (10)$$

$$x_{t-1} = \tau(\widetilde{x}_t^{ts}) \qquad (11)$$

In the THD, the guidance of data styles for different frequency components is more thoroughly represented and executed in the inference process. The style of the generated samples gradually aligns with the data styles from the real world. Therefore, the THD indicates which data styles in the generated samples originate from the real samples, making the inference process more interpretable. Furthermore, the THD reduces the interference of noise, thereby increasing the focus on crucial features in the inference process. As a result, the distributional bias between data generated by the diffusion model framework based on style-guided kernels and real data is also reduced. Algorithm 1 summarizes a single denoising time step of the THD under the diffusion model framework based on style-guided kernels.

## IV. EXPERIMENTS

### A. Experiment Setup

*1) Dataset*

The DS-Diffusion was tested and validated using four multivariate time series datasets from different domains: Stocks, Energy, MuJoCo, and Sine. The Stocks, Energy, and MuJoCo datasets came from real-world scenarios across different domains. The Sine dataset was created using a sine function, providing an ideal benchmark for comparison with other real-world datasets. The Stocks dataset comprised daily Google

---

**Algorithm 1** THD

**Input**: $\mathcal{D}^{L \times F} = \{x | x_0^{l \times F}, x_1^{l \times F}, \ldots, x_n^{l \times F}\}$, $x_t$, $t \in [0, T]$, $S_{tr}, S_{seas}$
**Parameter** : $\mathcal{T}_{tr}, \mathcal{T}_{seas}, \Upsilon$
**Output**: $x_{t-1}$

1: **while** $t > 0$ and $t < T$ **do**
2:     $x_t \leftarrow \Upsilon(x_t)$
3:     $x_t^{ts} \leftarrow \tau^{-1}(x_t)$
4:     $C_{tr}, C_{seas}, C_{error} \leftarrow \delta(x_t^{ts})$
5:     $\widetilde{C}_{tr} \leftarrow \mathcal{T}_{tr}(C_{tr}, S_{tr}, t)$
6:     $\widetilde{C}_{seas} \leftarrow \mathcal{T}_{seas}(C_{seas}, S_{seas}, t)$
7:     $\widetilde{x}_t^{ts} \leftarrow \widetilde{C}_{tr} + \widetilde{C}_{seas} + C_{error}$
8:     $x_{t-1} \leftarrow \tau(\widetilde{x}_t^{ts})$
9: **end while**
10: **return** $x_{t-1}$

---

stock data and included 6 variables: highest price, lowest price, opening price, closing price, adjusted closing price, and trading volume. The Energy dataset originated from the energy sector and consisted of continuous measurements with 28 variables that detailed the energy variations of the equipment [39]. MuJoCo was derived from the robotics field and generated time series data using a high-performance simulation engine for physical simulation and dynamic control, describing realistic physical behaviors of robots through 14 variables [40]. The Sine dataset contained 5 variables. Each variable exhibited distinct trends and periodicity. In the experiments, all variables in the datasets were generated and validated.

*2) Baseline*

The DS-Diffusion was meticulously compared with state-of-the-art time series generation models. ImagenTime [14] was used as the main baseline. The models compared, in addition to ImagenTime, included KoVAE [29], DiffTime [41], GT-GAN [42], TimeGAN [43], RCGAN [44], C-RNN-GAN [45], T-Forcing [46], P-Forcing [47], WaveNet [48], WaveGAN [49], and LS4 [30]. To ensure fairness and authenticity in the comparisons, the same experimental settings as those in [14] were employed.

*3) Hyperparameter Configuration*

The diffusion generative backbone of the DS-Diffusion (EDM) had 18 sampling steps. To demonstrate the performance improvement of the DS-Diffusion through rigorous comparison, the hyperparameters of the U-net used for denoising remained consistent with those in [14]. To obtain a trained diffusion model for the training and inference of the DS-Diffusion, the EDM was trained for 1000 epochs using the AdamW optimizer. In the style-guided kernels, the transformer models in the THD were trained with the Adam optimizer and were used to guide the denoising of different frequency components in the samples. The hyperparameters used in all experiments were described in detail in Table 1.

*4) Evaluation Metric*

The DS-Diffusion was evaluated in two parts. On one hand, the DS-Diffusion was assessed and compared with other advanced models using the same two metrics as in [14], including the predictive score (pred) and the discriminative score (disc). The pred was used to evaluate the quality of the generated samples, and the disc employed a discriminator to measure the similarity between the distributions of real data and generated data. On the other hand, to provide a more comprehensive evaluation of the distribution similarity between the generated data and the real data, Kullback-Leibler (KL) divergence, Jensen-Shannon (JS) divergence were used as evaluation metrics. To better quantify the evaluation of the DS-Diffusion, in addition to KL divergence and JS divergence, the Wasserstein distance (Wass) and the Kolmogorov-Smirnov test (KS) were additionally included to quantify the similarity of the probability density distributions between the generated data and the real data.

TABLE I
HYPERPARAMETERS IN EXPERIMENTS.

|  | Stocks | Energy | MuJoCo | Sine |
|---|---|---|---|---|
| **General** | | | | |
| image size | 8×8 | 8×8 | 8×8 | 8×8 |
| learning rate | $10^{-4}$ | $10^{-4}$ | $10^{-4}$ | $10^{-4}$ |
| batch size | 128 | 128 | 128 | 128 |
| **DE** | | | | |
| embedding | 8 | 8 | 8 | 8 |
| delay | 3 | 3 | 3 | 3 |
| **Diffusion** | | | | |
| U-net channels | 128 | 128 | 64 | 128 |
| in channels | [1,2,2,2] | [1,2,2,4] | [1,2,2,2] | [1,2,2,2] |
| sampling steps | 18 | 18 | 18 | 18 |
| **STL** | | | | |
| period | 24 | 24 | 24 | 24 |
| robust | True | True | True | True |
| **Transformer** | | | | |
| input dim | 13 | 57 | 29 | 11 |
| output dim | 6 | 28 | 14 | 5 |
| layers | 2 | 2 | 2 | 2 |
| embedding dim | 64 | 64 | 64 | 64 |
| learning rate | $10^{-3}$ | $10^{-3}$ | $10^{-3}$ | $10^{-3}$ |
| batch size | 128 | 128 | 128 | 128 |

### B. Comparison with Existing Methods

*1) Quantitative Comparisons.*

Quantitative comparisons of various generative models was first shown on four public datasets: Stocks, Energy, MuJoCo, and Sine. Table 2 illustrated the pred and disc metrics for DS-Diffusion, ImagenTime, and other advanced generative models on the real-world datasets: Stocks, Energy, and MuJoCo. ImagenTime served as a baseline and represented the most advanced results apart from DS-Diffusion. The experimental results in Table 1 indicated that DS-Diffusion achieved more advanced results compared to ImagenTime on the majority of datasets. On the three real-world datasets, DS-Diffusion reduced the average pred score and disc score of ImagenTime by 5.56% and 61.55%, respectively. The slight decrease in the pred score indicated an improvement in the quality of the generated samples, which aligned more closely with the objective physical laws of the real world. Additionally, the significant reduction in the disc score indicated that the distributional bias between the generated data and the real data was reduced. The similarity between the data generated by DS-Diffusion and the real data improved, bringing their data distributions closer together. The pred score and disc score for DS-Diffusion, ImagenTime, and other advanced generative models on the Sine dataset also validated the experimental results. Because the Sine dataset was generated from a sine function, it possessed a clear mathematical structure. This dataset was therefore used for testing and validating the performance of generative models in isolation. The experimental results in Table 2 indicated that DS-Diffusion reduced the pred score and disc score of ImagenTime by 7.09% and 94.42% on the Sine dataset, respectively. Additionally, the pred score and disc score of DS-Diffusion were closest to the state-of-the-art results. The results on the Sine dataset further validated the improvement in data quality and the reduction in distributional bias of the generated data.

Furthermore, the DS-Diffusion was compared with the state-of-the-art approach ImagenTime (baseline) on more metrics to validate the improvement of the proposed method over ImagenTime. The ImagenTime was denoted as "Latest", the DS-Diffusion was denoted as "Ours". The KL divergence, JS divergence, Wass distance, KS statistic between the generated data and the real data were presented in Table 3. Specifically, to mitigate the impact of randomness on the experimental results, the four metrics for the generative models on each dataset were calculated as the average values after multiple denoising inferences. The experimental results in Table 2 indicated that, compared to ImagenTime, the KL divergence and JS divergence between the generated data and the real data were reduced by 5.94% and 74.14%, respectively. The reduction in KL divergence and JS divergence further indicated that the distribution of generated data by the DS-Diffusion was statistically closer to that of the real data. This also validated the decrease in the disc score, which reflected the reduction in distributional bias. On average, the Wass distance was reduced by 96.37%, and the KS statistic was also reduced by 88.12%. The results of the additional quantitative comparisons further indicated that, compared to the ImagenTime, the DS-Diffusion reduced the distributional bias between the generated data and the real data in the inference process.

The computational efficiency of the proposed method was also worth attention. The training wall-clock runtime (WCR) and model size in terms of parameters (MP) were used to compare the computational efficiency of several state-of-the-art generative models. Using the commonly shared stocks and energy datasets [14], the efficiency-related metrics under the basic models' framework were presented in Table 4. The experimental results indicated that by increasing the model size in terms of parameters, the proposed method better captured the underlying patterns in the data and improved the quality of generated data. Particularly when high-performance computing resources were available, the increase in memory usage was acceptable as long as the generation quality was enhanced. Therefore, the proposed method demonstrated scalability in time series generation tasks. Since the data generation process did not participate in downstream classification or regression tasks, the increased computational resources for the generative model did not affect the efficiency of the target tasks.

Quantitative comparisons demonstrated that the diffusion model framework based on style-guided kernels and the THD effectively reduced the distributional bias between the generated data and the real data, and also enhanced the quality of the generated samples. Meanwhile, the inference process was more interpretable due to the guidance of data styles in the THD.

TABLE II
PRED SCORE AND DISC SCORE ON FOUR PUBLIC DATASETS (BOLD INDICATES BEST PERFORMANCE).

| Method | Stocks disc↓ | Stocks pred↓ | Energy disc↓ | Energy pred↓ | MuJoCo disc↓ | MuJoCo pred↓ | Sine disc↓ | Sine pred↓ |
|---|---|---|---|---|---|---|---|---|
| KoVAE | .009 ± .006 | .037 ± .000 | .143 ± .011 | .251 ± .000 | .076 ± .017 | .038 ± .002 | **.005 ± .003** | **.093 ± .000** |
| DiffTime | .050 ± .017 | .038 ± .001 | .101 ± .019 | .250 ± .003 | .059 ± .009 | .042 ± .000 | .013 ± .006 | **.093 ± .000** |
| GT-GAN | .077 ± .031 | .040 ± .000 | .221 ± .068 | .312 ± .002 | .245 ± .029 | .055 ± .000 | .012 ± .014 | .097 ± .000 |
| TimeGAN | .102 ± .021 | .038 ± .001 | .236 ± .012 | .273 ± .004 | .409 ± .028 | .082 ± .006 | .011 ± .008 | .093 ± .019 |
| RCGAN | .196 ± .027 | .040 ± .001 | .336 ± .017 | .292 ± .004 | .436 ± .012 | .081 ± .003 | .022 ± .007 | .097 ± .001 |
| C-RNN-GAN | .399 ± .028 | .038 ± .000 | .449 ± .001 | .483 ± .005 | .412 ± .095 | .055 ± .004 | .229 ± .040 | .127 ± .004 |
| T-Forcing | .226 ± .035 | .038 ± .001 | .483 ± .004 | .315 ± .005 | .499 ± .000 | .142 ± .014 | - | - |
| P-Forcing | .257 ± .026 | .043 ± .001 | .412 ± .006 | .303 ± .005 | .500 ± .000 | .102 ± .013 | - | - |
| WaveNet | .232 ± .028 | .042 ± .001 | .397 ± .010 | .311 ± .006 | .385 ± .025 | .333 ± .004 | .158 ± .011 | .117 ± .008 |
| WaveGAN | .217 ± .022 | .041 ± .001 | .363 ± .012 | .307 ± .007 | .357 ± .017 | .324 ± .006 | .227 ± .013 | .134 ± .013 |
| LS4 | .199 ± .065 | .068 ± .013 | .474 ± .003 | .251 ± .000 | .333 ± .029 | .062 ± .006 | .342 ± .007 | .132 ± .011 |
| ImagenTime | .037 ± .006 | **.036 ± .000** | .040 ± .004 | .250 ± .000 | .007 ± .005 | .033 ± .001 | .014 ± .009 | .094 ± .000 |
| DS-Diffusion | **.007 ± .006** | .036 ± .009 | **.010 ± .008** | **.240 ± .010** | **.005 ± .004** | **.030 ± .003** | .006 ± .005 | .093 ± .001 |

TABLE III
KL DIVERGENCE, JS DIVERGENCE, WASS DISTANCE, AND KS STATISTIC ON THE DATASETS (BOLD INDICATES BEST PERFORMANCE).

| Datasets | Metrics | Latest | **Ours** |
|---|---|---|---|
| Stocks | KL↓ | 1.0607 | **1.0302** |
| | JS↓ | 0.2355 | **0.0892** |
| | Wass↓ | 0.0185 | **0.0008** |
| | KS↓ | 0.0690 | **0.0130** |
| Energy | KL↓ | 1.8725 | **1.8228** |
| | JS↓ | 0.0866 | **0.0156** |
| | Wass↓ | 0.0217 | **0.0001** |
| | KS↓ | 0.1370 | **0.0060** |
| MuJoCo | KL↓ | 1.1781 | **1.0468** |
| | JS↓ | 0.1401 | **0.0588** |
| | Wass↓ | **0.0012** | 0.0014 |
| | KS↓ | 0.7228 | **0.0310** |
| Sine | KL↓ | 1.1939 | **1.1093** |
| | JS↓ | 0.3279 | **0.0183** |
| | Wass↓ | 0.0082 | **0.0005** |
| | KS↓ | 0.1850 | **0.0370** |

TABLE IV
COMPUTATIONAL RESOURCES IN TERMS OF WCR IN MINUTES (M) OR HOURS (H), AND MP IN MILLIONS (M).

| Method | | Stocks WCR | Stocks MP | Energy WCR | Energy MP |
|---|---|---|---|---|---|
| TimeGAN | | 2h 59m | 48K | 3h 37m | 1M |
| GT-GAN | | 12h 20m | 41K | 10h 39m | 57K |
| DiffTime | | 52m | 240K | - | - |
| LS4 | | 5h 30m | 2.7M | 2h | 2.1M |
| ImagenTime | | 1h 10m | 575K | 1h | 2M |
| Ours | Backbone | 1h 10m | 575K | 1h | 2M |
| | $\mathcal{T}_{tr}$ | 7m | 2.1M | 9m | 2.3M |
| | $\mathcal{T}_{seas}$ | 7m | 2.1M | 9m | 2.3M |
| | Total | 1h 24m | 4.8M | 1h 18m | 6.6M |

*2) Qualitative Comparisons.*

Qualitative comparisons provided an intuitive understanding of the generated samples, thereby better reflecting the performance of the generative model and the similarity between generated samples and real samples. In the assessment of the DS-Diffusion, three qualitative comparisons were employed on the four datasets. The comparisons included principal component analysis (PCA), t-distributed stochastic neighbor embedding (t-SNE), and visualization of the generated data. Furthermore, to enhance the credibility of the results, generated samples from the state-of-the-art ImagenTime were also visualized alongside the DS-Diffusion.

Firstly, PCA was used to visualize the distribution of generated data and real data in a two-dimensional space. By observing the distribution of different data in the PCA plot, the clustering and separation of generated data relative to real data could be intuitively understood. Fig. 2 displayed the PCA dimensionality reduction results for data generated by the DS-Diffusion, data generated by the ImagenTime, and real data on the four datasets. The experimental results indicated that the data generated by the DS-Diffusion exhibited a similar distribution to the real data in the PCA space, suggesting that the DS-Diffusion performed well in capturing the underlying structure of the data.

Secondly, t-SNE was employed for further visualization of the generated data and real data. As a nonlinear dimensionality reduction method, t-SNE was particularly suited for visualizing multivariate datasets. Unlike PCA, t-SNE better preserved local structures, bringing similar samples closer in the embedded space. By applying t-SNE to both generated data and real data, similarities and differences between the samples could be observed. Fig. 3 presented the t-SNE dimensionality reduction results for data generated by the DS-Diffusion, data generated by the ImagenTime, and real data on the four datasets. The t-SNE visualization results provided strong support for the performance of the DS-Diffusion.

Thirdly, the data generated by the DS-Diffusion was visualized in the form of probability density functions. The real data and the data generated by ImagenTime were also visualized to facilitate an intuitive assessment of the quality of the generated data. Fig. 4 presented the visualizations of both the generated data and the real data in the three real-world datasets. Similar to the results obtained from PCA and t-SNE, the data generated by the DS-Diffusion was found to be closer to the real data. Furthermore, compared to ImagenTime, the trends in the data generated by DS-Diffusion aligned more closely with the trends observed in the real data.

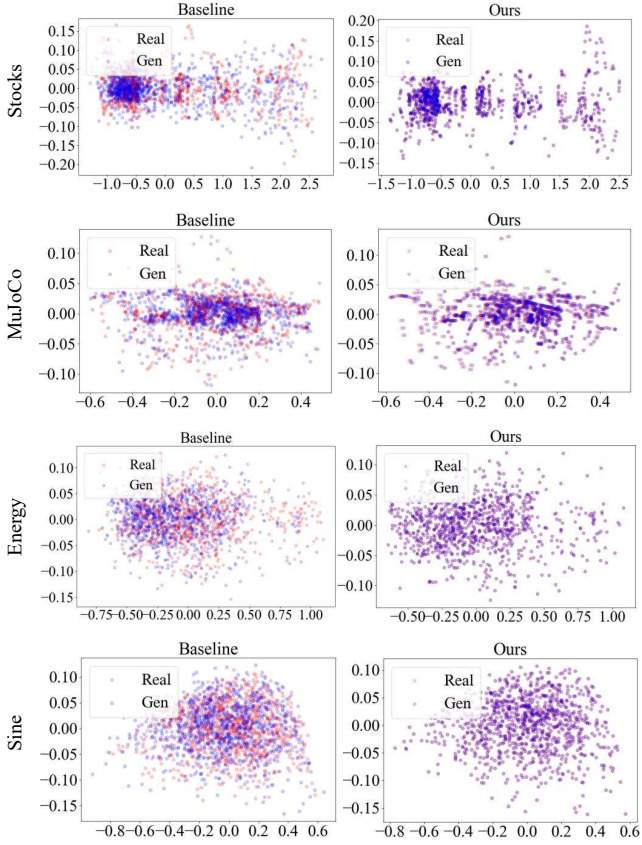

Fig. 2. Qualitative comparisons of generated data and real data with PCA.

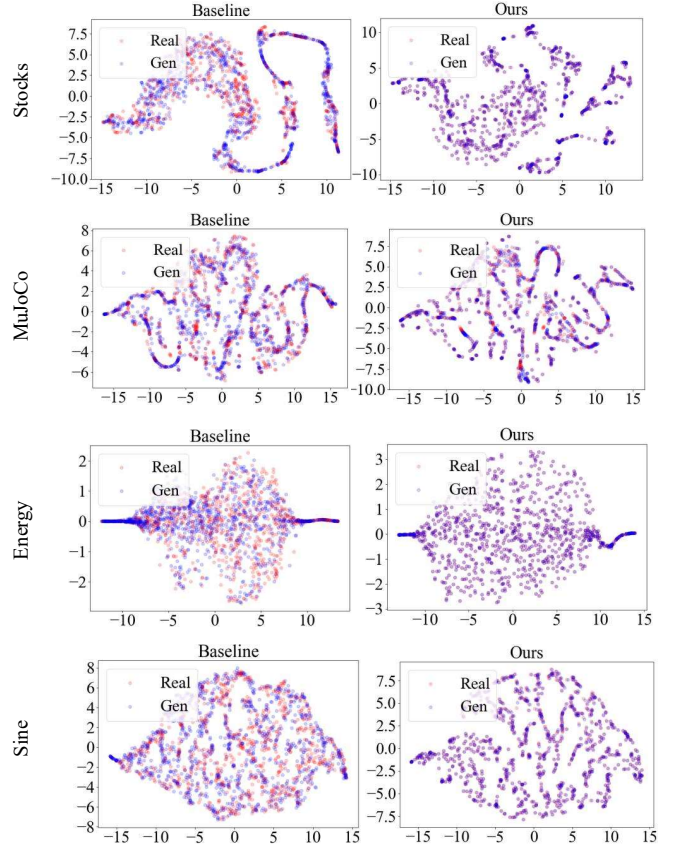

Fig. 3. Qualitative comparisons of generated data and real data with t-SNE.

Finally, the data generated by the DS-Diffusion (Gererated) in the latent space was compared with the real data (Ground truth). The generated data was randomly selected from different datasets and variables. In multiple generation results, the shaded areas represented the standard deviation bands of the generated data. The uncertainty intervals of the generated data were indicated by the standard deviation bands, reflecting the range of fluctuations. The visual comparison results provided a more intuitive evaluation. Fig. 5 showed the comparison between the real data and the generated data from the real-world datasets. The comparisons spanned different variables and time points. Fig. 5 offered a more intuitive validation of the results from PCA, t-SNE, and probability density functions. The DS-Diffusion was able to generate data that conformed to real physical laws by leveraging data styles guidance. The deviation between the generated data and the real data was minimal, making the generated data sufficiently indistinguishable from the real data. Additionally, the comparative results in Fig. 5 further demonstrated the realistic data generation capabilities of the DS-Diffusion. The DS-Diffusion was also extendable to time series generation and imputation tasks.

The results of qualitative comparisons indicated that the distributions of data generated by the DS-Diffusion more closely overlapped with the distributions of real data under the qualitative comparisons. Therefore, the data generated by the DS-Diffusion exhibited a higher similarity to the real data. The distributional bias between the generated data and the real data was further reduced. Qualitative comparisons offered a multidimensional perspective for assessing the performance of generative models and the quality of generated data. The visual results from qualitative comparisons also validated the experimental metrics in quantitative comparisons. Additionally, the combination of qualitative comparisons and quantitative comparisons enhanced the credibility of the results, achieving an effective complementarity between quantitative data and qualitative insights.

### C. Ablation Study

To further evaluate the contributions of each method and component, an ablation study was conducted. The selection of STL and Transformer was validated independently. To verify the contribution of STL, commonly used time series analysis methods such as fourier transform (FT), continuous wavelet transform (CWT), and time-domain gradients (TG) were employed to extract features that guided data generation.

TABLE V
ABLATION STUDIES(BOLD INDICATES BEST PERFORMANCE).

| Target | Remain | Stocks disc↓ | Stocks pred↓ | Energy disc↓ | Energy pred↓ | MuJoCo disc↓ | MuJoCo pred↓ | Sine disc↓ | Sine pred↓ |
|---|---|---|---|---|---|---|---|---|---|
| - | DS-Diffusion | **.007 ± .006** | **.036 ± .009** | **.010 ± .008** | **.024 ± .010** | **.005 ± .004** | **.030 ± .003** | .006 ± .005 | **.093 ± .001** |
| STL | FT | .019 ± .004 | .038 ± .005 | .030 ± .008 | .026 ± .040 | .012 ± .005 | .035 ± .000 | **.006 ± .004** | .094 ± .000 |
|  | CWT | **.007 ± .006** | .037 ± .008 | .017 ± .010 | .025 ± .020 | .010± .004 | .033 ± .005 | .008 ± .005 | .094 ± .001 |
|  | TG | .048 ± .013 | .038 ± .006 | .036 ± .016 | .028 ± .020 | .016 ± .002 | .036 ± .003 | .021 ± .013 | .095 ± .001 |
| Style-guided Kernels | - | .037 ± .006 | .036 ± .000 | .040 ± .004 | .025 ± .000 | .007 ± .005 | .033 ± .001 | .014 ± .009 | .094 ± .000 |
|  | GRU | .005 ± .004 | .036 ± .009 | .203 ± .035 | .027 ± .020 | .018 ± .008 | .032 ± .005 | .008 ± .002 | .094 ± .000 |
|  | MLP | .054 ± .018 | .040 ± .010 | .276 ± .044 | .032 ± .017 | .019 ± .002 | .037 ± .014 | .018 ± .010 | .096 ± .025 |
| THD | THD ($\mathcal{T}_{tr}$) | .124 ± .019 | .039 ± .001 | .220 ± .010 | .251 ± .000 | .225 ± .016 | .039 ± .002 | .066 ± .006 | .098 ± .002 |
|  | THD ($\mathcal{T}_{seas}$) | .023 ± .020 | .036 ± .010 | .191 ± .015 | .251 ± .000 | .017 ± .010 | .031 ± .002 | .006 ± .005 | .094 ± .001 |

To evaluate the contribution of the transformers, widely used time series processing models such as GRUs and MLPs were utilized to replace the transformers within the diffusion model framework. Furthermore, the contributions of style-guided kernels and the THD were validated independently. Specifically, $\mathcal{T}tr$ and $\mathcal{T}seas$ in the THD were added separately to assess their contributions. The main metrics of the ablation experiments were described in Table 5.

Firstly, in terms of method selection, experimental results demonstrated the necessity of choosing STL and Transformer. STL effectively separated trends and seasonality without requiring prior assumptions about frequency, making it suitable for non-stationary time series. Its robustness to outliers and flexibility in modeling complex patterns rendered it particularly advantageous in various real-world applications. FT was more suitable for handling datasets with clear periodicity, demonstrating optimal performance on the standard periodic sine dataset. But it performed worse than STL on datasets with weak or less obvious periodicity. The guiding effect of CWT was similar to that of STL. Although CWT was capable of simultaneously capturing both time-domain and frequency-domain information, its decomposition results depended on the selected mother wavelet and scale parameters, resulting in poorer stability during data generation. The advantage of TG lay in its ability to characterize local changes within the sequence, but it lacked the capability to separate long-term trends and periodic patterns. In contrast, STL robustly separated trend and seasonal components, , thereby outperforming FT, CWT, and TG in DS-Diffusion generation tasks. Moreover, transformers exhibited greater advantages over GRUs and MLPs in time series modeling due to its self-attention mechanism, which captured global dependencies and multi-scale features simultaneously, thereby enhancing long sequence modeling capability and representation accuracy.

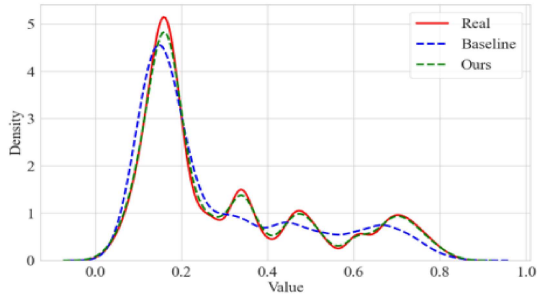
(a) Stocks: Generated Data and Real Data.

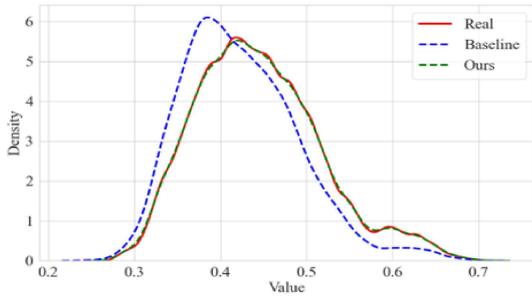
(b) Energy: Generated Data and Real Data.

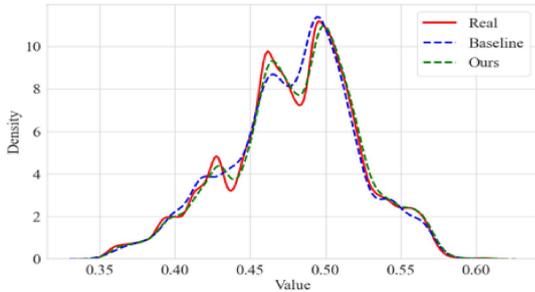
(c) MuJoCo: Generated Data and Real Data.

Fig. 4. Visualization of Generated Data and Real Data in Real-world Datasets.

Secondly, in terms of the contributions of components, experimental results indicated that in the style-guided kernels, the $\mathcal{T}tr$ played a dominant role in the inference process, and the $\mathcal{T}seas$ served an auxiliary role. The $\mathcal{T}tr$ established a sequence profile that aligned with the characteristics of the real samples, thereby allowing the generated samples to better reflect the features of the real samples. Meanwhile, the $\mathcal{T}seas$ played a crucial role in capturing the inherent periodicity of data styles, particularly in maintaining the stylistic consistency and distribution characteristics between generated data and real data. For instance, in the MujoCo dataset, when the $\mathcal{T}tr$ was ablated, the disc score and pred score increased to (0.225, 0.039), representing a 4400% and 30% increase compared to the DS-Diffusion (0.005, 0.030), respectively. The substantial rise in the disc score indicated an increase in the distributional bias between the generated data and the real data after the removal of the $\mathcal{T}tr$. The similarity of the generated sample profiles to those of the real samples significantly decreased. Additionally, when the $\mathcal{T}seas$ was ablated, the disc score and pred score rose to (0.017, 0.031), reflecting a 240% and 3.33% increase compared to the DS-Diffusion (0.005, 0.030). While the metrics showed varying degrees of improvement after the ablation of the $\mathcal{T}seas$, the increases were less pronounced than those observed with the ablation of the $\mathcal{T}tr$. The stylistic consistency between the generated data and the real data was affected. The experimental results confirmed the auxiliary role of $\mathcal{T}seas$ in the style-guided kernels. The variations in metrics across the other three datasets exhibited similar trends to those seen in the MujoCo dataset. Therefore, the ablation study demonstrated the complementary roles of the $\mathcal{T}tr$ and the $\mathcal{T}seas$ in the style-guided kernels. The critical contributions of the $\mathcal{T}tr$ and the $\mathcal{T}seas$ in enhancing the quality of generated samples and reducing distributional bias were highlighted. The $\mathcal{T}tr$ and the $\mathcal{T}seas$ were complementary and indispensable. The ablation study provided a clear illustration of the contributions of each component. The specific contributions of the style-guided kernels to generation tasks were also showcased, particularly highlighting the THD. Furthermore, the ablation study provides strong support for the design of the DS-Diffusion and lays the groundwork for future research.

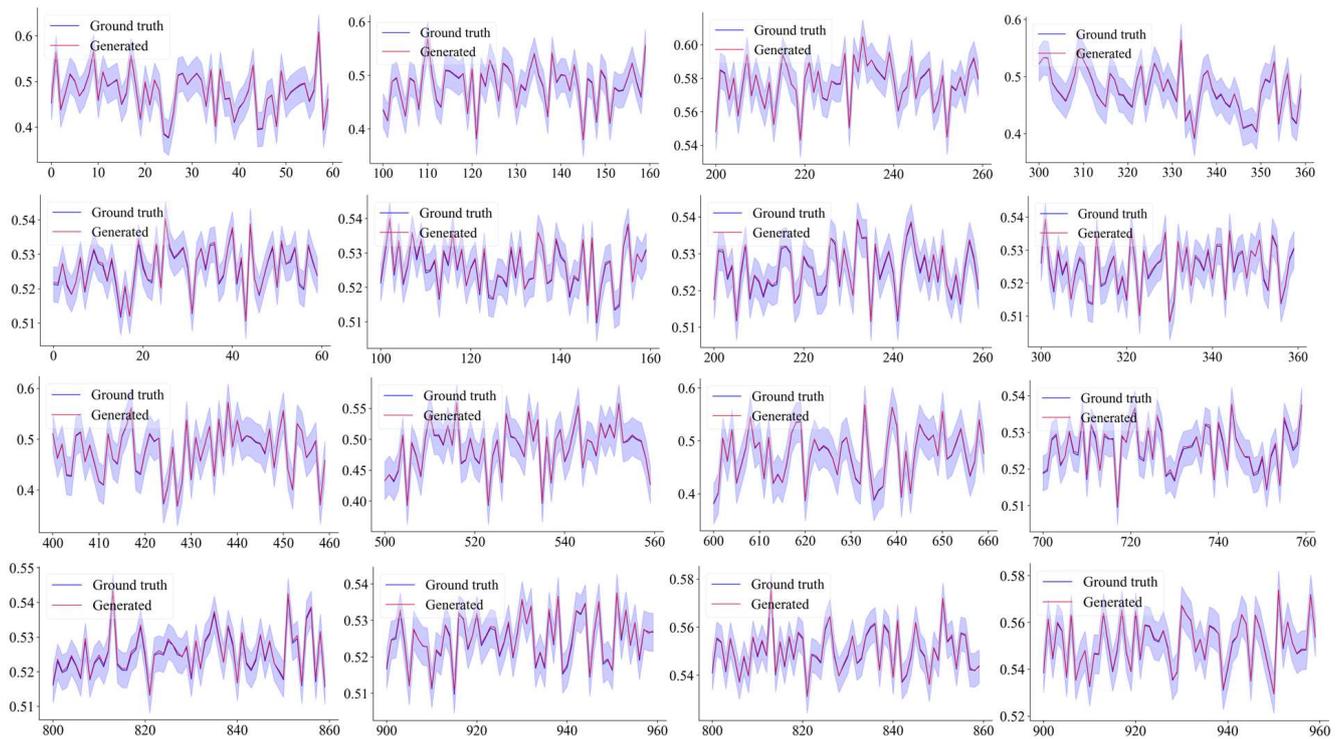

Fig. 5. Comparisons: Real Data (Ground truth) and Generated Data (Gererated) from Real-world Datasets.

## V. CONCLUSION

In the DS-Diffusion, a diffusion model framework based on style-guided kernels is proposed to enhance the model's flexibility and adaptability for specific scenarios without the need for retraining to introduce new conditions. The THD is proposed to guide the denoising of components at different frequencies in the inference process. As a result, the distributional bias between the generated data and real data is reduced, and the inference process is more interpretable. The proposed DS-Diffusion can serve as a foundational model for generating samples according to different data styles.

In future research, the proposed method could be extended to applications in agriculture, food, and healthcare. For instance, in agriculture, olfactory sensors are utilized for quality detection of agricultural products such as rice, vegetables, and fruits. The DS-Diffusion could be employed to augment olfactory sensor data, thereby enhancing the quality detection capabilities of downstream models. In the healthcare sector, blood glucose monitoring data is widely used for the health management of diabetes patients. The DS-Diffusion could be applied to augment blood glucose time series signals, improving the downstream models' ability to recognize fluctuations and abnormal states in blood glucose levels. Furthermore, the DS-Diffusion, with its capability for realistic data generation, could be applied to time series prediction and imputation tasks, and optimizing representation methods for time series at different frequencies presents a promising research direction. We also aim to enhance model computational efficiency through new style extraction methods, model structure optimization, and knowledge distillation.

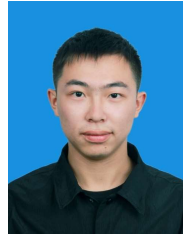
**Songyu Ding** received the B.S. degree from the School of Electronic Information, Wuhan University, Wuhan, China, in 2023, and the M.S. degree from the School of Electrical and Electronic Engineering, Nanyang Technological University, Singapore, in 2024.

He is currently pursuing the Ph.D. degree with Harbin Institute of Technology, Harbin, China. His current research interests include computer vision, generative models, and data augmentation.

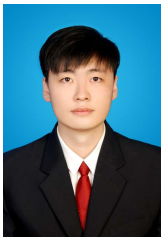
**Mingchun Sun** received the B.S. and M.S. degrees in University of Electronic Science and Technology of China, Chengdu, China, in 2018, and 2021, respectively.

He is currently pursuing the Ph.D. degree with Harbin Institute of Technology, Harbin, China. His current research interests include data augmentation and deep learning on datasets with limited samples.

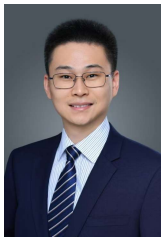
**Rongqiang zhao** received the Ph.D. degree in Harbin Institute of Technology, Harbin, China, in 2018.

He is currently an Associate Professor with the Faculty of Computing, Harbin Institute of Technology, Harbin 150001, China, and also with the National Key Laboratory of Smart Farm Technologies and Systems, Harbin Institute of Technology, Harbin 150001, China.

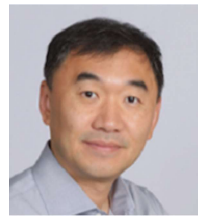
**Jie Liu** (Fellow, IEEE) received the Ph.D. degree in electrical engineering and computer science from the University of California at Berkeley, Berkeley, CA, USA, in 2001.

He is currently a Chair Professor with the Harbin Institute of Technology, Shenzhen, China. His research interests include artificial intelligence, control engineering, Internet of Things, and computer system.

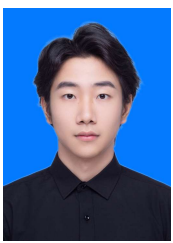
**Hengrui Hu** received the B.Eng. degree in Automation from the School of Future Technology, Harbin Institute of Technology, Harbin, China, in 2025.

He is currently pursuing the M.Sc. degree with the Faculty of Computing, Harbin Institute of Technology. His research interests include multimodal deep learning.